# Automatic Classification of Games using Support Vector Machine

Ismo Horppu[1], Antti Nikander[1], Elif Buyukcan[1], Jere Mäkiniemi[2], Amin Sorkhei[3], Frederick Ayala-Gómez[1]
[1]Rovio Entertainment, ({name}.{surname}@rovio.com),
[2]Zalando SE, (jere.makiniemi@zalando.com),
[3]Unity Technologies, (amin@unity3d.com)

*Abstract*—Game developers benefit from availability of custom game genres when doing game market analysis. This information can help them to spot opportunities in market and make them more successful in planning a new game. In this paper we find good classifier for predicting category of a game. Prediction is based on description and title of a game. We use 2443 iOS App Store games as data set to generate a document-term matrix. To reduce the curse of dimensionality we use Latent Semantic Indexing, which, reduces the term dimension to approximately 1/9. Support Vector Machine supervised learning model is fit to pre-processed data. Model parameters are optimized using grid search and 20-fold cross validation. Best model yields to 77% mean accuracy or roughly 70% accuracy with 95% confidence. Developed classifier has been used in-house to assist games market research.

## I. Introduction

CLASSIFICATION of games in custom genres is useful in market trend analysis and when planning to create a new game. Native categories available in app stores are not always detailed enough therefore multiple companies have developed their own genres and classifications (more about this in section II). It is important to look what are market trends with respect to game genres when a game developer is planning to create a new game. This information can be used to check for example saturation of markets (e.g. does it look possible to enter market with a specific genre game without too much competition) or to check if some genres are trending up or down. Saturation can be checked against number of similar genre games or against revenue or download share distribution by genre (is it dominated by few big games). Also, volume of daily new games is high therefore classification needs to be automated to save time.

Game classifications have been available online from a specialized website called GameClassification.com [16] and Apple RSS feed [17]. The RSS feed contained up to two native categories per game. GameClassification.com includes data such as gameplay, purpose, target audience and market for each game. The website is a spinoff from academic research project [1]–[2].

Apple's RSS feed data is automatic as it is up-to-date, whereas game classification web site categories are updated manually by users of the web site. We decided not to use the game classification web site as it does not have all iOS games. In addition, we were not aware if there exists easy data access such as RSS feed or an API. If we had been totally satisfied with Apple's native categories then we would have used them directly without need to develop classification model. However, we decided to specify our own categories that better fit our needs. As a result of this we had to develop method which can classify unseen game to a category using its description and title.

There are two ways to classify games: using supervised or unsupervised methods. We have tried both. Benefits of using supervised method is that one can get quite accurate result. However, disadvantage is that training data labels, genres, are not easy to get in volume – someone needs to manually create them (that is if you are not using previously mentioned online available categories). Secondly, you cannot find unseen genres which can limit market trend analysis. Unsupervised learning, such as Latent Dirichlet Allocation topic model, can solve the previously mentioned two issues. Yet, it comes with its own challenge - found genre (mixtures) may not be easy for a human to understand. In this paper we focus to a supervised learning method.

*Contribution*

Our contributions provide way to enable automatic prediction of custom game genres with good enough accuracy. To be specific, we contribute in i) developing methodology for predicting game genres and ii) showing empirical result.

*Organization*

Section II summarizes related work for classification of games, whereas section III describes creation of in-house categories and related work. Overview of the classification process is given in section IV. Data setup is summarized in section V, whereas modelling method is described in detail at section VI. Experiments details are given in section VII, and finally in Section VIII we present our conclusions.

## II. Related Work For Intelligent Classification

This paper is a revisited version of work done in 2013. When most part of work for this paper was done, we were not aware of other automatic and "intelligent" publicly described methods for classifying games. By intelligent we mean that classification is done using game description and title. There is an open-source tool called depressurizer and Android CatApplus application for classification, which were available in 2013 as well. Depressurizer is for Valve's Steam to categorize PC games. However, it uses pre-set store categories of games and is not for mobile games. CatApplus app, automatically categorizes applications using pre-set native categories. Both of these applications are not intelligent, although they are useful for end users.

Since 2013 there has been development in research of intelligently classifying games. Namely, Suatap and Patanukhom have applied convolutional neural networks to icon and screenshot data to classify games [21]. In addition, Zheng and Jiang used deep learning to predict game genres using text and image data [18]. Jiang also did Master Thesis about the topic [19]. Interestingly, Zheng's and Jiang's image data-based classification model did not perform as well as text data-based model. This is not totally unexpected finding: if you have ever looked game in-store screenshots then you may have found it difficult to classify some games based on them only (gameplay video would be better but modelling it is more complicated). Their combined multi-modal model did not perform much better than model using just text data.

### III. CREATION OF IN-HOUSE GAME CATEGORIES AND RELATED WORK

Many people have created taxonomies of games: a thorough study has been done by Wolf [3]. Gunn et al. [4] summarize that current genres are merely representational and "missing any form of delineation based on interactivity". Lewis et al. [5] provide mapping of the mental space of game genres. They state that it is challenging because of subjective evaluations and many axes on which games can vary. Klabbers [6] presents a framework for classifying games. Petrenz [7] assesses approaches to genre classification in his master thesis. Parts of his thesis apply to categorization of games.

More recently multiple companies have realized that custom (= not standard mobile application store categories) game genres and categories are useful, for market analysis for example. Multiple companies such as GameRefinery, App Annie, AppsFlyer and SensorTower have published their own taxonomies [22]-[26]. Some companies have gone to more details: GameAnalytics having genre-category-subcategory [27] and Newzoo having also theme-art style [26]. In addition, Verto Analytics has published a study about how demographics of players and time of playing differ between genres [20].

We wanted to use categories which are meaningful in our company, such as for game market research. Creation of categories took few iteration rounds: initial set of categories which we planned was not good. Set of categories evolved after some thorough thinking done by in-house experts. Final set consisted of 21 categories.

We want to state that categorization of games is subjective process: classifications done independently by two or more professionals may not always agree.

### IV. OVERVIEW OF CLASSIFICATION PROCESS

Fig. 1 depicts process flow for categorization of a game based on its description and title which both we got from Apple's App Store RSS feed. Our categorization method uses Support Vector Machine (SVM) supervised learning method.

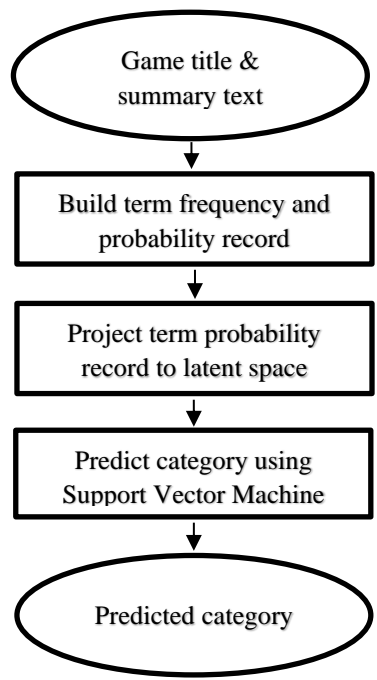

FIG. 1. FLOW FOR PREDICTION OF A GAME CATEGORY.

### V. DATA SETUP

#### A. Data Collection

We extracted game description and title data from Apple's App Store iOS RSS feed on July 2nd 2013, and we restricted our collection to games with English description in top 300 paid, free and grossing lists for iPhone and iPad from Australia, Canada, Germany, Spain, Finland, France, Great Britain, Italy, Sweden, Turkey and United States. Duplicates were removed.

Games from Apple's App Store RSS feed had to be re-categorized after creating our categories to build final data set. We re-categorized random sample of 2443 games from our collection because time available for categorization by experts was limited. Categorization of the games was done using our prior information, playing games and checking descriptions, screenshots and sometimes gameplay videos. We also decided to give only one main label for each game although we know that this is not optimal: a game can be a combination of multiple categories. However, this decision made modelling a bit simpler.

#### B. Data Pre-Processing

We carried data pre-processing using following steps:
1) **Removal of line feed, punctuation and special characters** (see Appendix for details).
2) **Removal of potentially 'misleading' company names** such as Big Fish. Namely, word fish may get connected to

fishing games. This processing was limited to few companies due to time constraints.

3) **Excluding other games advertisement from description texts**. For example, at the end of an Electronic Arts (EA) game description there was an advertisement "Don't miss our other exciting games! Scrabble, The Sims 3, Tetris, …". Due to time limitations this processing was done to such advertisements text used by six game companies: EA, Activision, NaturalMotion, PlayFirst, SEGA and Imperial Game Studio.
4) **Fry's [8] top 25 English words were removed** (see Appendix for list of words). We considered these words to be non-informative.
5) **Word stemming:** for example, conquered and conquering are changed to conquer.
6) We created **stemmed word vocabulary** from data set. This contained 15969 words.
7) **Noise terms were removed** based on amount of mutual information between terms and categories. This made vocabulary decrease to 5340 words and had negligible effect on classification accuracy.
8) Finally, we **reduced vocabulary** to 2400 latent dimensions.

We excluded higher order n-grams because of significantly higher computation time and memory requirements.

Fig 2. (a) represents word cloud of top 50 stemmed words. Top 51-100 stemmed words are displayed in the Fig 2. (b). Stemmed words are from initial vocabulary before term reductions were done. We have underlined words, based on our subjective judgment, which contain quite clear information about game categories. Top 50 words are not as informative for classification of games as top 51-100 words. This is expected when spotting non-informative words among the top 50 words such as www, http, twitter, Facebook, iPhone, iPad and HD.

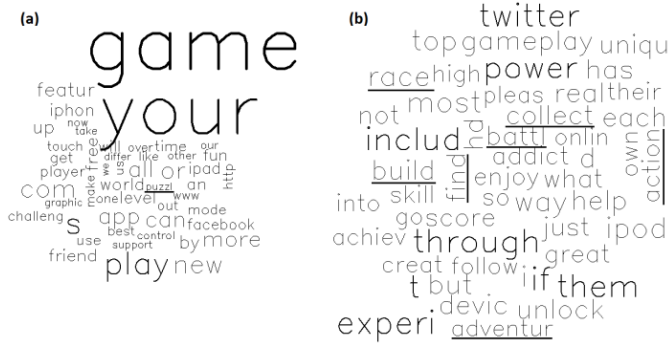

Fig 2. Word clouds for top 50 words (a) and top 51-100 words (b). We have underlined words which we think have information about categories.

### C. Creation of Term Frequency and Probability Matrices

Rows in document-term matrix are games from our data set. Columns are vocabulary terms. Cell j,i of a term frequency matrix (tfm), which is also known as document-term matrix, contains number of occurrences of term (stemmed word) j in document i. In our case document consists of game title and description.

At first term frequency matrix is filled by number of term occurrences. Then it is typically pre-processed. In this study we tested following pre-processing methods:
- **term probability matrix**: term frequencies are normalized by converting them into discrete probability distribution of terms in vocabulary. This is done by dividing term frequencies of a game by total sum of term frequencies for the game,
- **term frequency matrix itself**: using frequencies and
- **term boolean occurrence matrix**: 0/1 indicator for existence of a term in a document.

We noticed that the term probability matrix performed best, yielding to highest mean accuracy. We also tried few implementations of term frequency-inverse document frequency (tf-idf). However, those did not improve accuracy.

## VI. MODELLING METHOD

### A. Using Information Theory for Selection of Useful Terms

The vocabulary contains so-called noise terms. A noise term means terms which may be frequent but contain none or little information about any category. Before the latent space of terms is constructed, we remove the noise using an algorithm that looks for major variance directions. Without this kind of reduction noise terms would impact latent axes.

Information Theory has a suitable quantity for computing information between random variables X and Y. This quantity is called mutual information and it is defined as

$$I(X,Y) = \iint p(x,y) \log(\frac{p(x,y)}{p(x)p(y)}) \, dx \, dy, \quad (1)$$

where p(x,y) is the joint probability distribution function of X and Y, p(x) and p(y) are marginal probability distribution functions of X and Y and integrals are over domains of X and Y. If random variables are independently distributed, then (1) equals to 0.

In our case random variable X is term probability variable (a column from term probability matrix) and Y is indicator variable of a category. This means that outer integral over domain of Y changes to weighted summation over values 0 and 1. In addition, we approximate integral over domain of X by using 100 equally spaced bins. We used base 2 logarithm in (1) for interpretability because then information is measured in number of bits.

After computation of mutual information for all stemmed terms and categories we decided a term selection rule. We used rule that if maximum value of mutual information for a term is over a threshold value then we add term to filtered vocabulary. Maximum is calculated over all pairwise mutual information between the term and 21 categories. We had to trade between accuracy of model and its computational efficiency, which is controlled by dimension of vocabulary, to select a suitable thresholding value. With some test and try we found out that threshold value of 0.0035 bits provided good result.

Fig 3. demonstrates top 10 stemmed terms, with respect to mutual information, for one of our categories. It is obvious that these terms are not independent from each other. Selection criteria was not optimal but its implementation was straightforward and efficient.

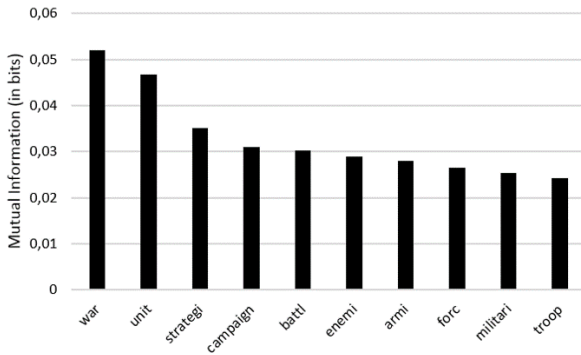

Fig. 3. Mutual information for top 10 stemmed words for one of our categories.

*B. Data Dimensionality Reduction*

We applied dimensionality reduction to document-term data to lighten computational complexity and reduce risk of over fitting. This was done by using latent semantic indexing (LSI), which itself uses singular value decomposition (SVD) method [30]. LSI is also known as latent semantic analysis (LSA) [31]. Purpose of LSI is to reduce term matrix dimension by finding linear combinations of vocabulary terms in game titles and descriptions. This method is often applied in text prediction tasks: Yang et al. [9] have used it. Dimensionality reduction is described in detail as next.

Let *A* be p x n matrix of term probabilities of p games with term vocabulary size n. In our case p equals to 2443 and n to 5340 (after non-informative terms have been removed). Matrix A can be written using singular value decomposition as

$$A = U * D * V^T \quad (2)$$

where *U* is p x n matrix with orthogonal columns, *D* is diagonal n x n matrix having singular values (which all are non-negative), $V^T$ is n x n orthogonal matrix, * is matrix multiply operator and $^T$ is matrix transpose operator. Column vectors of *U* are called left singular vectors of *A* and columns of *V* as right singular vectors.

Matrix *A* in (2) can be approximated as

$$A_k = U_k * D_k * V_k^T \quad (3)$$

where k highest singular components are used. Valid values for k are: $1 \leq k \leq \min(p,n)$.

The k-rank approximation can be used, by solving $U_k$ from (3), to project term probability vector of a game to k-dimensional latent space. As an example, a term probability vector b, which in our case is 5340 dimensional corresponding to all terms in vocabulary, is projected as

$$b' = b * V_k^T * D_k^{-1} \quad (4)$$

However, the right side of equation (4) requires inversing matrix $D_k$, which brings some additional computational challenges. To avoid the matrix inverse, we scale projected vector *b'* by energy levels of $D_k$ forming vector *b˜*. Projection (4) followed by scaling simplifies to

$$b˜ = [b * V_k^T * D_k^{-1}] * D_k = b * V_k^T \quad (5)$$

We found that this scaling improves our classification performance. Also, the right side of (5) is numerically more stable than (4) because it does not require matrix inverse.

*C. Classification learning method*

We tested three classification models: CART, naive Bayes, and SVM. The first two did not perform well enough (accuracies were between 40% and 60%). Support Vector Machine (SVM) is known to perform well in various classification problems [10]-[12]. We decided to focus our experiments using SVM as it clearly outperformed the other two model families.

Support vector machines were originally developed for pattern recognition [13] by Vapnik and Chervonenkis. Basic idea in SVM classification is to fit linear classifier to data set in augmented space. When a Gaussian (radial basis function) kernel is used, like in our case, then augmented space is infinite dimensional. Linear decision boundaries at augmented space correspond to non-linear decision boundaries at original space. Regularization of model is used to avoid over-fitting to data set.

Advantages of SVM are flexibility, via non-linear decision boundaries, and having mathematically nice optimization problem. Nice refers to fact that optimization problem is solvable using quadratic programming (QP). This means that converge to global optima is guaranteed. This is good feature compared for example to multilayer perceptron (MLP) neural networks which optimization can get stuck to local optima.

In SVM classification context a so-called soft-margin optimization problem with radial kernel function has regularization constant C and gamma parameter. Constant C sets the relative importance of minimizing so called amount of slack and maximizing margin. Gamma parameter controls how far influence a single sample has. Low gamma value means far and high value "close". In other words, inverse of gamma is proportional to variance parameter used with Gaussian function.

Schölkopf et al. [14] have modified SVM optimization problem by introducing nu-parameter which eliminates need for constant C which has unbounded non-negative range. They have named this method as nu-SVM. Nu parameter has limited range (0, d] where d is problem dependent but at maximum 1 (see [15] for details). Nu represents an upper bound on the fraction of training samples which are incorrectly predicted and a lower bound on the fraction of support vectors.

VII. EXPERIMENTS

*Experimental setting*

We did all data handling such as pre-processing, predictions and evaluation using R. We loaded App Store RSS feed to CSV file using a Java program which we implemented. We used qdap (to remove top 25 English words) and Rstem (word stemming) R packages for preprocessing. We ran discretize2D and mi.empirical functions, from the entropy package, to reduce the dimensionality of vocabulary using mutual information criteria. Finally, for classification we used SVM nu-classification implementation from e1071 package.

Computer used for experimenting was Intel i5-3247U mobile processor (2 cores, 1.8GHz base frequency / 2.8GHz turbo frequency, 8GB RAM) running Windows 7. Script was executed on single core. Running time for model search, see details about it in next section, was around five days. Script was executed on single core.

*Model Search and Validation*

We searched for good SVM model using grid search. We did some initial model searches which indicated that good classification performance is with latent dimensionality k value 2400. Grid search was done over SVM complexity gamma parameter values {0.01, 0.5, 1, 1.5, 2, 2.5, 3, 3.5, 3.75, 4, 4.5, 5, 7.5, 10, 12.5, 15, 22.5, 25, 30, 35, 37.5, 45, 128, 256} and nu parameter values {0.005, 0.01, 0.0125, 0.015, 0.0175, 0.02, 0.025, 0.03, 0.035, 0.04, 0.045, 0.05, 0.07, 0.08, 0.10, 0.15, 0.25}. The previous parameters corresponded to 408 different models.

We could not use higher than 0.25 value for nu parameter because such values resulted into infeasible nu parameter error during nu-SVM model fitting (see [15] for details on valid nu range). This happens if data classes are unbalanced as they are in our case. We tried two balancing methods to expand nu parameter range. First applied method was balance boost which increases proportion of rare classes in data set. Second tried method was use of inverse class weights. Neither of tried balancing methods improved result - therefore we used nu-parameter range (0, 0.25].

We used cross validation to select model: use one out of 20 data chunks for testing and rest 19 chunks for training and calculate test set classification accuracy. We repeated this so that all data chunks were used as testing. Finally, we computed mean accuracy as average over all test accuracies computed from 20 test sets.

Fig. 4 represents mean accuracy of all the 408 models as function of number of support vectors. Remark that number of support vectors is function of training data, gamma and nu parameters. Over-fitting of data starts when number of support vectors exceeds approximately value 2250. Mean accuracy starts to decrease when over-fitting begins.

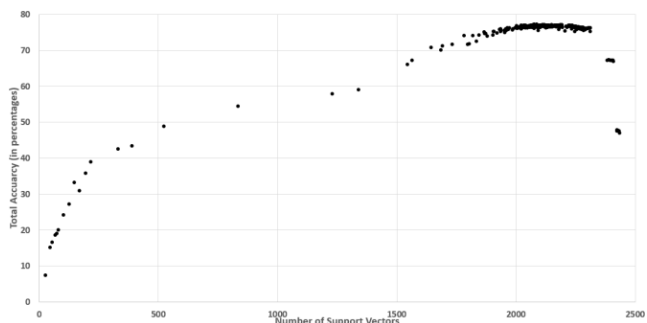

Fig. 4. Number of support vectors versus mean accuracy.

We checked information of term probability matrix preserved by latent space. Fig. 5 depicts singular values of term probability matrix. Singular values reflect amount of variability in terms of game description and title term data captured by latent space axes. They are sorted in descending order - from most important axis to least important axis. According to the figure it looks that variability has been captured well.

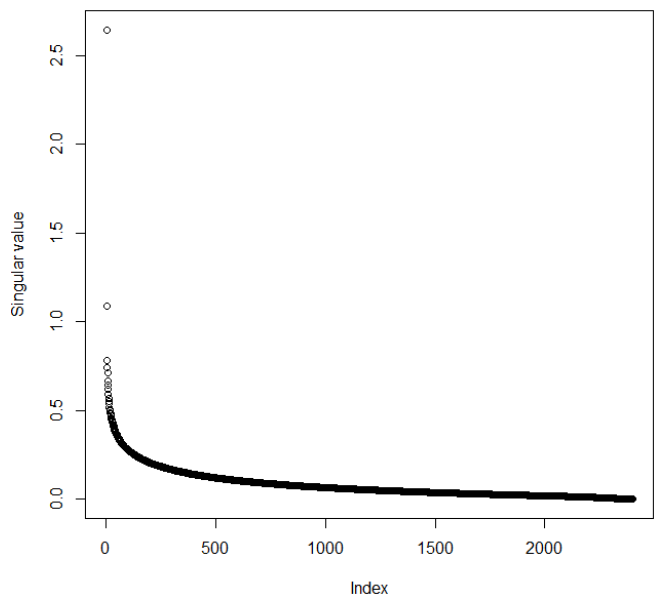

Fig. 5. Singular values of term probability matrix.

The grid search found best model with parameters combination k=2400, gamma=3.5 and nu=0.025. Best model uses 2074 support vectors.

The best model uses 2074 support vectors out of approximately 2320 (= 19/20 blocks * 2443 records) training samples what equals to 89% of training samples. This may mean that there is not much regularity in the data. We know that typically one aims for as low fraction of support vectors as possible with reasonable classification performance. However, it is expected that text classification with multiple classes, such as in our case, creates complicated decision boundaries which require high percentage of training samples for good performance.

*Results*

Accuracy of our best model was 77.4 percentages with standard deviation 3.6. Mean accuracy is approximately Gaussian random variable which means that with 95% confidence we have about 70% accuracy.

State of the art methods by Jiang and Cheng [18] and Suatap and Patanukhom [21] have achieved 47-50% top-1 accuracy with 17 classes. Our result is also for top-1 accuracy. However, these two other results are not comparable with our result because of few reasons. Firstly, our sample size is greatly less ~2400 versus Jiang and Cheng's 50000 and Suatap and Patanukhom's 24308. Secondly, Jiang & Cheng's original data set has multiple genres per game, and they select one randomly to make it single class prediction task. Thirdly, class balances are likely differently. Fourthly, our data is from year 2016 and their six years newer. Finally, used classes are different: we use custom classes defined and labeled by our in-house experts, whereas they use app store native classes.

We did also simple visual quality analysis of classifications by checking locations of classification successes and failures against locations of training data. This was done for best model and data was randomly split into 70:30 into training and test data sets. Fig. 6 (a) depicts location of training data set samples. Coordinates are first two major latent axes, X is most important axis and Y is second most important axis, of latent space presentation of term probability matrix. Fig. 6 (b) represents test data set samples in the two major axes. Fig. 6 (a) shows that games are located almost in one circular cluster when considering only two major latent space axes.

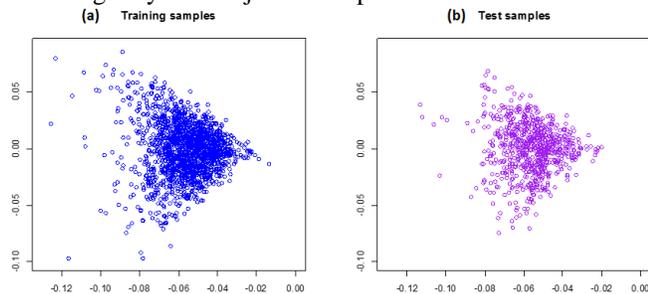

Fig. 6. Training samples (a) and test samples (b). X-axis is $1^{st}$ and Y-axis is $2^{nd}$ latent axis of term probability matrix.

Locations of correctly and incorrectly classified samples are depicted in Fig. 7 (a) and Fig. 7 (b). Comparing those to locations of all test samples, see Figure 6 (b), we notice that locations of correctly and incorrectly classified test records are randomly distributed among distribution of samples.

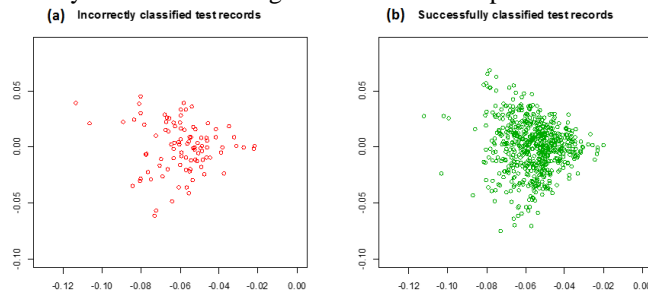

Fig. 7. correct predictions (a) and incorrect predictions (b).

## VIII. CONCLUSION

This paper investigates the task of classifying games based on the title and description using a custom classification. Our proposed approach is to reduce the noise terms using LSI and then classifying the genres using SVM. Based on empirical experimentation, the proposed classification yields enough accuracy for the task of market trend analysis, and such approach may be useful for people in gaming industry.

Open future research lines include studying:
- using native (for example AppStore) categories to assist in predicting custom categories when applicable,
- using other classification methods (xgboost, chirp),
- replacing mutual information in variable selection by global correlation coefficient (which has well-defined scale),
- using human created classification rules to assist predictions done by a model,
- expanding term vocabulary (including selected bi-grams and perhaps higher order n-grams),
- better than latent semantic indexing (LSI) method for dimensionality reduction,
- computationally more efficient SVM algorithm for doing predictions (perhaps using LaSVM that allows online learning too [28] or GPU implementation [29]),
- various class balancing methods and
- using non-English text (asian markets, romance languages).

ACKNOWLEDGMENT

The authors want to thank following people: Tero Raij, Jami Laes, Janne Koivunen, Touko Herranen, Joonas Kekki, Jarno Martikainen, Assel Mukhametzhanova, Miikka Luotio and Otso Virtanen. All these people have been supportive and helped in development of this paper. In addition, special thanks to Jussi Huittinen and Michail Katkoff: they planted seeds, perhaps without knowing, to make revisiting this paper happen.

## APPENDIX

*A. List of removed characters:*

The following characters were removed: ?, *, _, @, -, +, !, =, ®, ™, •, …, ", ", ~, &, —, ., ,, #, ', ', (, ), -, ;, ©, :, 0, 1, 2, 3, 4, 5, 6, 7, 8, 9, /

*B. List of Fry's top 25 English words*

Following words were removed: the, of, and, a, to, in, is, you, that, it, he, was, for, on, are, as, with, his, they, I, at, be, this, have, from.